\documentclass[journal]{IEEEtran}  % Comment this line out if you need a4paper

\IEEEoverridecommandlockouts                              % This command is only needed if 
\usepackage{graphics} % for pdf, bitmapped graphics files
\usepackage{epsfig} % for postscript graphics files
\usepackage{mathptmx} % assumes new font selection scheme installed
\usepackage{times} % assumes new font selection scheme installed
\usepackage{amsmath} % assumes amsmath package installed
\usepackage{amssymb}  % assumes amsmath package installed
\usepackage{subfig}
\usepackage{xcolor} % Add this line to load the xcolor package
\usepackage{color,soul}
\usepackage{graphicx}
\usepackage{xy}
\usepackage{atbegshi}

\usepackage{import,bm}
\newcommand{\mt}[1]{\bm{#1}}

\title{Deep-Learning Estimation of Weight Distribution Using Joint Kinematics for Lower-Limb Exoskeleton Control 
}

\author{Clément Lhoste, Emek Barış Küçüktabak, Lorenzo Vianello, Lorenzo Amato, Matthew R. Short,\\ Kevin~M.~Lynch, and Jose L. Pons% <-this % stops a space
\thanks{C. Lhoste and L. Vianello are with the Legs and Walking Lab, Shirley Ryan AbilityLab, Chicago, IL, USA.}
\thanks{E. B. Küçüktabak is with the Center for Robotics and Biosystems, Department of Mechanical Engineering, Northwestern University, Evanston, IL, USA and Shirley Ryan AbilityLab.}
\thanks{L. Amato is with the Shirley Ryan AbilityLab and the Biorobotics Institute, Scuola Superiore Sant’Anna, 56025 Pontedera, Italy and Department of Excellence in Robotics \& AI, Scuola Superiore Sant'Anna, 56127 Pisa, Italy.}
\thanks{M. R. Short is with the Shirley Ryan AbilityLab and Department of Biomedical Engineering, Northwestern University, Evanston, IL, USA}
\thanks{K. M. Lynch is with the Center for Robotics and Biosystems and Department of Mechanical Engineering, Northwestern University.}
\thanks{J. L. Pons is with the Shirley Ryan AbilityLab, Center for Robotics and Biosystems, Department of Mechanical Engineering, Department of Biomedical Engineering, and Department of Physical Medicine and Rehabilitation, Northwestern University.}

}

\begin{document}

\AtBeginShipoutNext{\AtBeginShipoutUpperLeft{%
  \put(\dimexpr\paperwidth-1cm\relax,-1.5cm){\makebox[0pt][r]{\framebox{Accepted for publication in the IEEE Transactions on Medical Robotics and Bionics (TMRB).}}}%
}}

\maketitle
\thispagestyle{empty}
\pagestyle{empty}

%%%%%%%%%%%%%%%%%%%%%%%%%%%%%%%%%%%%%%%%%%%%%%%%%%%%%%%%%%%%%%%%%%%%%%%%%%%%%%%%
\begin{abstract}

In the control of lower-limb exoskeletons with feet, the phase in the gait cycle can be identified by monitoring the weight distribution at the feet. This phase information can be used in the exoskeleton's controller to compensate the dynamics of the exoskeleton and to assign impedance parameters.
Typically the weight distribution is calculated using data from sensors such as treadmill force plates or insole force sensors. However, these solutions increase both the setup complexity and cost.
For this reason, we propose a deep-learning approach that uses a short time window of joint kinematics to predict the weight distribution of an exoskeleton in real time. The model was trained on treadmill walking data from six users wearing a four-degree-of-freedom exoskeleton and tested in real time on three different users wearing the same device. This test set includes two users not present in the training set to demonstrate the model's ability to generalize across individuals. Results show that the proposed method is able to fit the actual weight distribution with $R^2=0.9$ and is suitable for real-time control with prediction times less than 1 ms. Experiments in closed-loop exoskeleton control show that deep-learning-based weight distribution estimation can be used to replace force sensors in overground and treadmill walking. 


\end{abstract}

%%%%%%%%%%%%%%%%%%%%%%%%%%%%%%%%%%%%%%%%%%%%%%%%%%%%%%%%%%%%%%%%%%%%%%%%%%%%%%%%
\section{Introduction}

Lower-limb exoskeletons are increasingly being utilized in everyday tasks and rehabilitation for individuals with gait impairments. In clinical settings, these exoskeletons are typically classified into two main types: partial assistance and full mobilization. Full mobilization exoskeletons are designed for patients with severe motor control disorders, providing autonomous movement of the legs regardless of the patient's input~\cite{Baud2021}. Conversely, partial assistance exoskeletons aid in the patient's movement while still requiring their active input.
In this paper, we focus on partial assistance exoskeletons, as they are commonly employed in rehabilitation. Depending on the functional level of the patient, these exoskeletons use techniques like haptic guidance/assistance~\cite{deMiguelFernndez2023} or error augmentation/resistance~\cite{9513580} to facilitate the (re)learning of walking behaviors.
% This approach, commonly employed in rehabilitation~\cite{Siviy2022}, often involves the use of haptic guidance~\cite{deMiguelFernndez2023}. Several other techniques are employed to facilitate the learning of motor tasks through haptic disturbances. These include methods like haptic error augmentation or resistive training~\cite{9513580}.
Implementing both haptic assistance and resistance strategies requires precise control over the interaction torques between the user and the exoskeleton, ensuring effective and safe physical interaction.

\begin{figure}[t]
    \centering
    \includegraphics[width=0.9\linewidth]{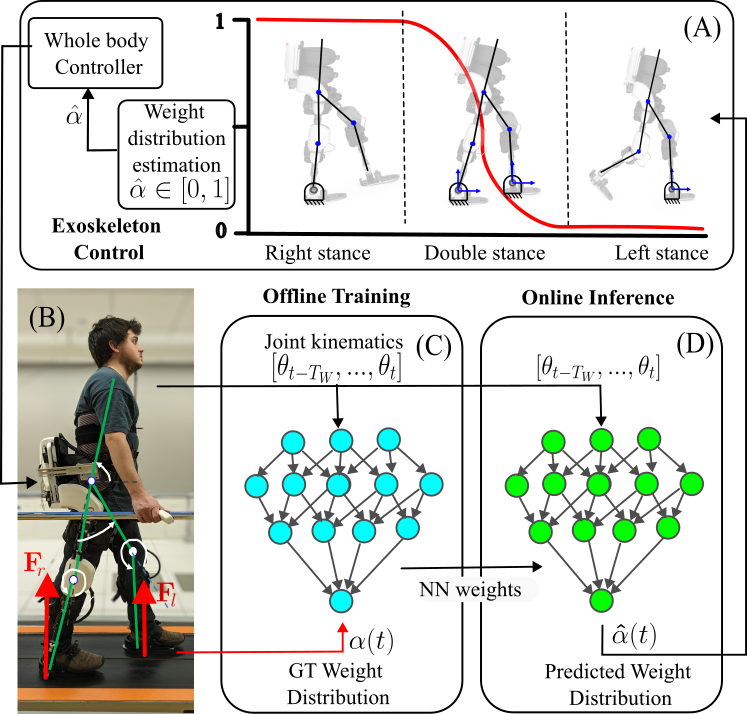}
    \caption{\small Weight-distribution regression: a deep-learning approach is used to estimate the weight distribution ($\alpha$) of a user-exoskeleton couple in real-time and to control the robot accordingly. Part (A) and (B) of the schema, Sec.\ref{sec:method_exoskeleton}, provides an overview of the exoskeleton hardware and controller as well as the calculation of $\alpha$ using force sensors applied to the feet and used as ground truth value. Part (C), Sec. \ref{sec:method_offline}, presents the structure of the deep-learning model and its offline training using a time-window (300ms) of kinematics information as input and ground truth values of $\alpha$ as output. Part (D), Sec \ref{sec:method_online}, displays how the trained model is used in the closed-loop control to predict the weight distribution ($\hat{\alpha}$) using kinematic information. The estimated values are passed in real-time (inference time~$<1$ms) to the exoskeleton controller.} 
    \label{fig:concept}
    \vspace{-0.5cm}
\end{figure}

% % An example population requiring assistive devices and rehabilitation is Spinal cord injuries (SCIs), affecting 250,000-500,000 patients globally each year~\cite{SCIBOOK}.
% The modulation of interaction forces between humans and exoskeletons commonly follows a top-down approach. This approach is typically structured into three hierarchical levels: high-level, mid-level, and low-level~\cite{Baud2021, Kim2022}. The highest level includes the calculation of desired interaction forces based on the activity performed (such as overground walking, stairs, ramps).
% % 3 control levels
% %Interaction force control of an exoskeleton can be divided into three hierarchical levels. The highest level includes the calculation of desired interaction forces based on the activity performed.
% % The various activities performed are usually organized in a high-level controller.
% The mid-level controller is often used to estimate the different states within an activity (e.g. gait state during walking) and compensating the dynamics of the system. 
% % This hierarchy allows to properly compensate for the dynamics of the system in any given situation (for each activity and each gait state).
% Finally, the low-level controller computes motor commands based on the current state and interaction force error.

Typically, the control of the interaction between a user and exoskeleton adopts a hierarchy of control levels: high-level, mid-level, and low-level~\cite{Baud2021, Kim2022}. For lower-limb exoskeletons, the high-level controller calculates the desired interaction torques tailored to specific ambulatory activities, such as walking on flat terrain, stairs or ramps. The role of the mid-level controller is to estimate the various states within an activity, for example, identifying the gait states (e.g., swing and stance) during walking and setting a desired interaction torque behavior accordingly. Finally, the low-level controller is responsible for compensating the exoskeleton's dynamics and generating motor commands, based on the desired interaction torque profile identified by the mid-level controller.

In this three-level approach, an accurate, real-time estimation of the gait state in the mid-level and low-level is essential for properly controlling the exoskeleton. 
% This value is used to assign the interaction force profiles in the mid-level controller and to compensate for the full body weight in the low-level controller. 
%Detection of gait events like heel strike and toe-off is based on measuring ground reaction forces, joint positions, or velocities~\cite{8386443, 8487913}. These gait events are typically integrated into a state machine framework to facilitate switching between distinct gait states: left stance, right stance, and double stance~\cite{Baud2021}. However, achieving a smooth and seamless transition between these states is vital. 
A common approach to gait state detection is to use gait events (e.g., heel strike, toe-off) to identify the switching between discrete gait states: left stance, right stance, and double stance~\cite{8386443, 8487913, Baud2021}.
However, achieving a smooth transition between these states is vital to ensure a control input that limits unexpected behaviors for the user.
Küçüktabak et al.~\cite{2024haptic} proposed a whole-exoskeleton closed-loop compensation (WECC) controller that uses the ratio of ground reaction forces (GRF) (i.e., weight distribution) to approximate the double stance phase as a transition between the left and right stance.
The weight distribution, quantified as the stance interpolation factor $\alpha$, smoothly changes from zero (left stance) to one (right stance) during double stance, as visualized in Fig.~\ref{fig:concept}.
Nonetheless, the requirement of additional sensors for GRF measurements, such as treadmill force plates or insoles, increases the complexity and cost of these systems.

Machine learning models have demonstrated their ability to estimate gait states using only joint kinematics of humans walking without exoskeletons, thus eliminating the need for GRF measurements~\cite{MUNDT202029, Karatsidis2016, https://doi.org/10.1111/cgf.14635}. 
% However, these studies focus on humans alone without exoskeleton. Moreover, the implementation of these approaches in real-time faces hurdles, primarily due to the extended processing time required by the motion capture system and the impracticality associated with managing multiple IMUs. 
% For these reasons, continuously estimating the gait state in real-time using standard sensors on an exoskeleton presents a significant challenge.
Motivated by these results, recent studies have proposed real-time estimation of gait states with lower-limb exoskeletons. % in real-time without the use of GRF sensors. %using standard sensors on an exoskeleton. % presents a significant challenge.
% Fewer sensors require less calibration and maintenance of the robot, resulting in a simpler experience for both patients and therapists.
 %Jung et al. \cite{s151127738} used a classification approach to detect stance and swing, using sensors available on a lower-limb exoskeleton. Specifically, the authors used sagittal plane orientations of the thigh and shank, and their angular velocities. This classification will result in non-continuous torque modeling.
%(multilayer perceptron and nonlinear autoregressive with external inputs (NARX)). sensors = 
%Liu et al. \cite{Liu2016} used joint angular sensors of an exoskeleton to classify the gait state (among 8 possibilities). This approach was user-independent; however, the discrete approach would again result in torque or modeling discontinuities.
For instance, Jung et al.~\cite{s151127738} and Liu et al.~\cite{Liu2016} used data from joint encoders and an inertial measurement unit (IMU) positioned on the trunk to classify the gait cycle during walking. 
% These sensors consist of encoders in both approaches, in addition to IMU positioned on the trunk in~\cite{s151127738}.
These studies divided gait cycles into two (i.e., stance and swing) and eight gait states, respectively. %The latter approach was user-independent.
However, discrete control states result in torque or modeling discontinuities.
Camardella et al.~\cite{9415159} and Lippi et al.~\cite{icinco21} address this issue by implementing a linear regression and a Neural Network model for regression with joint angles as inputs. The model's output smoothly transitions between left and right stance, providing a continuous representation of the gait state. 
Despite this advancement, these approaches present some limitations. First, they only consider the present joint configuration, discarding past information. This approach might lead to inaccuracies in predictions as the weight distribution is influenced not only by the current posture, but also by the velocity and acceleration of the motion.
% We believe this could result in less accurate predictions because the weight distribution is influenced not only by the current posture but also by how the movement is executed, particularly where speed and acceleration play a crucial role.
Moreover, few considerations in these studies are given to the generalization across users and the prediction time performances. The latter is essential for real-time implementation of the exoskeleton's whole-body controller. 

To address the aforementioned limitations, we present a deep-learning approach for estimating the weight distribution of a user wearing a lower-limb exoskeleton. This approach utilizes a short time window of kinematic data extracted from sensors integrated in the exoskeleton. For the validation of predictions, we compared the proposed approach with the method in \cite{icinco21} which only uses the instantaneous posture.
% this approach with using as input solely the current posture as in \cite{icinco21} showing that a short-term time window can improve the prediction accuracy.
Employing distinct users for training and testing resulted in a user-independent estimation, eliminating the need for a calibration phase. The proposed deep-learning method was trained on walking data from six users, and its closed-loop performance was evaluated on three users (two of which were not included in the training set) during treadmill and overground walking. The results showed that weight distribution can be accurately estimated using kinematic data with an $R^2$ of 0.9 averaged across users. In addition, this approach was compatible with real-time requirements, achieving a prediction time less than 1 ms.
%Furthermore, this estimation is demonstrated to be independent of both user variability and movement speed.

% We introduce an approach using deep learning to replace the mid-level controller of the exoskeleton and eliminate the need for footplate sensors or sensorized treadmills. Weight distribution is estimated using kinematic data in real-time, in a user- and speed-independent approach. This method removes the sensor calibration phase and reduces development and maintenance costs.

\section{Methods}

\subsection{Exoskeleton}
\label{sec:method_exoskeleton}
We validated our proposed method on a commercially available lower-limb exoskeleton (ExoMotus-X2, Fourier Intelligence, Singapore). The exoskeleton system has four active degrees of freedom (DoF) at the hip and knee, and two passive DoFs at the ankle (Fig. \ref{fig:concept}B). %The system can record joint positions, joint velocities, and states of an integrated key-fob. 
The exoskeleton was modified to include strain gauges to estimate joint torques, and an Inertial Measurement Unit (IMU) on the backpack to measure the orientation of the trunk~\cite{2024haptic}. 

%explain the sensor !
In this study, to measure GRFs, and therefore calculate weight distribution, two distinct devices were employed: a sensorized, split-belt treadmill and force-sensitive resistor (FSR) footplates. The treadmill is equipped with eight 3-DoF force plates (9047B, Kistler), offering highly accurate readings and a simple calibration process to zero the measured forces. However, its use restricts the practical applications of the exoskeleton and prevents its integration into overground scenarios.
To address this limitation, FSR footplates were installed beneath the soles of the exoskeleton, facilitating the measurement of forces between the user-exoskeleton couple and the ground. Each footplate is equipped with 16 force-sensitive resistors, amounting to a total of 32 sensors across both footplates. These sensors interface with the ground through a rigid aluminum sole and rubber bearings, adding a degree of compliance. Although these sensors are more complex to calibrate and set up in comparison to the treadmill, they allow the exoskeleton to be used during overground walking.

% weight distribution 
GRFs enable the calculation of weight distribution in the user-exoskeleton couple. This allows an accurate and smooth transition between the left and right stance dynamic models of the exoskeleton (Fig. \ref{fig:concept}A). In this work, we calculated the weight distribution as the ratio of the vertical GRF of each foot similar to \cite{2024haptic},
\begin{equation}
    \alpha = \frac{{F_{\text{right},y}}}{{F_{\text{left},y}} + {F_{\text{right},y}}},
    \label{eq:alphainmotion}
\end{equation}
where $\alpha \in [0,1]$ is the \emph{stance interpolation factor}, while $F_{\text{left},y}$ and $F_{\text{right},y}$ represent the vertical GRFs.

Communication between motors and sensors is achieved using the CANOpen protocol. The controller is implemented on a ROS and C++ based open-source platform called the CANOpen Robot Controller (CORC) \cite{10.1007/978-3-030-69547-7_47}.

\subsection{Weight Distribution via Deep-Learning Regression}
\label{sec:method_offline}

In an effort to simplify the equipment mounted on the exoskeleton, we propose a deep-learning approach to provide an estimate $\hat{\alpha}$ of the stance interpolation factor using only kinematic information as input.
Our approach is based on a Long Short-Term Memory (LSTM) network that uses a time window of kinematic values ($x_t = [\mt{\theta}_{t-T_W}, ..., \mt{\theta}_{t-t_1}, \mt{\theta}_t]$) as input and returns the current $y_t = \hat{\alpha}_t$ as an output. 
The vector ($\mt{\theta}$) is composed of the joint angles of the hip and knee joints for both legs, and the backpack angle in the sagittal plane ($\mt{\theta} = [\theta_{l,\_hip}, \theta_{r,\_hip}, \theta_{l,\_knee}, \theta_{r,\_knee}, \theta_{b}]$). 
To evaluate the effects of short-term history, models with a time window of $T_W = 300$~ms (considering the recommendations from \cite{s20216345}) were compared with models $T_W=0$~ms that only use the instantaneous kinematic posture at time $t$ (similar to \cite{icinco21}).

\begin{figure}[t]
\centering
\vspace{0.3cm}
\includegraphics[width=0.99\linewidth]{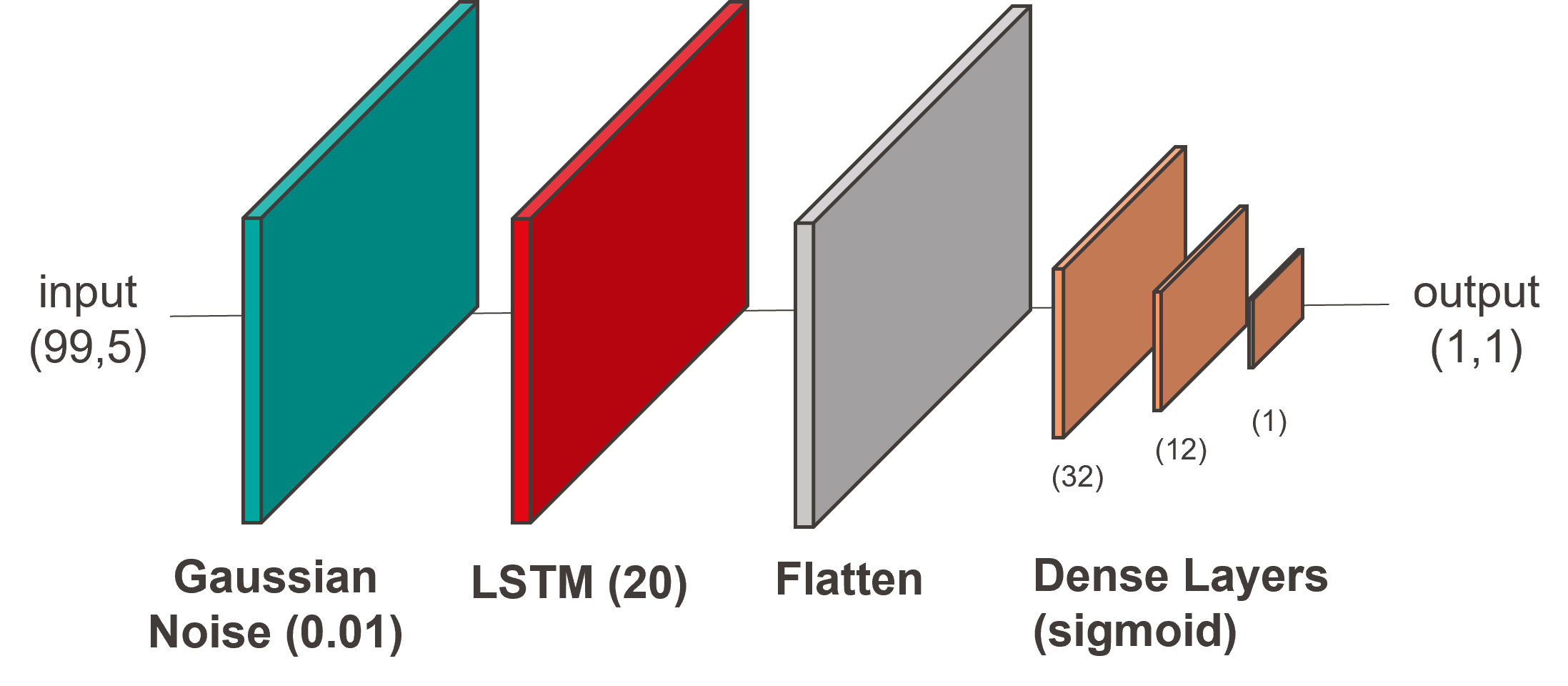}
\caption{\small Model architecture to predict the stance interpolation factor (parameters are shown in parentheses). Gaussian noise is used to corrupt the data and improve generalization. The LSTM layer extracts features in the temporal domain of the input. Dense layers serve as regression layers, transforming the features into the desired output. The input vector contains the last 300~ms kinematic information ($[\mt{\theta}_{t-T_W}, ..., \mt{\theta}_{t}]$) collected with a frequency of 333~Hz, resulting in an input vector of size $(99,5)$. The output of the network is the stance interpolation factor ($\alpha_t$) of size $(1, 1)$ at time $t$.}
\label{stance-interp-model}
\vspace{-0.3cm}
\end{figure}

The architecture of the LSTM model is similar to the architecture in~\cite{9842329} and is presented in Fig. \ref{stance-interp-model}. The sensor data was sampled with a frequency of 333~Hz resulting in an input matrix of size $(99,5)$ (99 time instances of five kinematic parameters) for the short-term history model and an output data of size $(1,1)$. Gaussian noise (standard deviation of 0.01) is used in the first layer to introduce variability to the input data ($x_t$), allowing better generalization. The LSTM layer allows the model to learn dependencies between the different time steps in the window of data. This layer is configured with 20 units. Finally, dense layers transform the features extracted from the LSTM to a single value representing the weight distribution ($\hat{\alpha}_t$). A sigmoid activation function is used in these layers to always return a value between $0$ and $1$ in accordance with the definition of the \emph{stance interpolation factor}; this provides a usable output from the network as an input to the control framework. Hyperparameters were selected using a heuristic approach. %We used K-fold cross-validation and selected parameters that improve the performance metrics selected: Mean Square Error (MSE) and Coefficient of determination ($R^2$). 
We employed leave-one-out cross-validation and used the coefficient of determination~($R^2$) and Mean Square Error~(MSE) between the predicted~($\hat{\alpha}$) and actual~($\alpha$) stance interpolation values as performance metrics to evaluate the chosen parameters.%We reported the mean ($\pm$ standard deviation) of the coefficient of determination~($R^2$) and Mean Square Error~(MSE) between the predicted~($\hat{\alpha}$) and actual~($\alpha$) stance interpolation values.
% selected parameters that enhance the chosen performance metrics, specifically, Mean Square Error (MSE) and Coefficient of Determination ($R^2$).

The model was trained using the ADAM optimizer~\cite{kingma2014adam}. We utilized the MSE loss function to measure and minimize the discrepancy between the predicted ($\hat{\alpha}$) and actual ($\alpha$) values obtained from treadmill measurements. The training process was conducted over 10 epochs, processing the data in batches of 256 samples, to allow efficient weight adjustments while ensuring comprehensive coverage of the data.
%The training of the model utilized the ADAM optimizer, with mean square error loss, during 10 epochs and with a batch size of 256.
The LSTM model was trained using the TensorFlow library (v2.15.0, Google); we employed TensorFlow Lite (v.2.13.0, Google) to make predictions in real time. The converted lite model is called by a Python script that runs in parallel to CORC in C++. The communication between these two components is achieved via ROS.
%The two components communicate using ROS: one Subscriber to receive the data from the X2 and create the input windows and one Publisher to send the Stance Interpolation Factor to the X2.

\subsection{Experimentation}
\label{sec:method_online}

We used data collected in two of our previous studies \cite{2024haptic, Kucuktabak2023virtual} to select the parameters of the LSTM model trained offline (Fig. \ref{fig:concept}C) and then to validate its usability in real-time applications (Fig. \ref{fig:concept}D). The training data includes six users wearing the exoskeleton and walking for a total of 210 minutes. Users walked on a treadmill with a speed of 0.3~m/s either in the haptic transparent mode (i.e., zero desired interaction torques) or in a haptic rendering mode where virtual rotational spring-damper elements were rendered at their hip and knee joints~\cite{2024haptic}. The  rendered virtual impedance has stiffness and damping values of 30 Nm and 6 Nms, and neutral position of 25\textdegree and -45\textdegree at hip and knee, respectively.
After the offline training and validation of prediction capability, we evaluated the capability of the LSTM network for closed-loop predictions of the stance interpolation factor $\hat{\alpha}$ in two experiments. 

In the first experiment, three users wore the exoskeleton and walked for one minute on a sensorized treadmill for each mode (i.e., haptic transparency and rendering). The stance interpolation factor for the exoskeleton controller was calculated either using the treadmill force plates as ground truth ($\alpha$), or estimated using the LSTM model ($\hat{\alpha}$). The measured or estimated value was used as an input to the WECC controller for interaction torque estimation and closed-loop compensation control in real-time~\cite{2024haptic}. 
% Fig. \ref{fig:concept} displays the framework proposed in this work.
The performance of the proposed model was assessed by comparing the spatiotemporal gait features and resultant interaction torque errors. Reported and visualized interaction torque errors in the plots are calculated using the ground truth $\alpha$ obtained from the treadmill force plates, and divided by each user's weight. We report all performance values as the mean $\pm$ standard deviation across users and steps unless otherwise specified. 
To assess the network's ability to generalize to unobserved walking conditions and different controllers, we evaluated the model on three speeds (0.14 m/s, 0.25 m/s, and 0.47 m/s) and two exoskeleton control modes (haptic transparency and rendering). Moreover, two of the three test users were not in the training set. Inter-user variability is removed from the presented box plots by demeaning the data of each condition/user pair and subsequently adding the overall mean of each condition, including three users.
% To compare the average interaction error averaged over the three users, recorded data was demeaned for each condition/subject pair and the overall mean of each condition, including three subjects was subsequently added.

In the second experiment, two users who were not featured in the training set performed trials of overground walking. The deep-learning approach was trained on a dataset of healthy individuals walking on a treadmill, thus we aim to demonstrate the model's ability to generalize to more naturalistic walking behaviors. In this experiment, $\alpha$ was calculated using the FSR footplates, while $\hat{\alpha}$ was calculated using the deep-learning approach. The users walked for 10~meters at a self-selected speed, repeated three times for each condition; interaction torques and resultant joint angles were used to assess the performance.

% In the second experiment, two users did ten-meter overground walking three times for each condition. This time the stance interpolation factor $\alpha$ was calculated using the FSR footplates, while $\hat{\alpha}$ was calculated using the same deep-learning approach. 
% The user self-selected the speed and interaction torques are used to assess the performance. The deep-learning approach has been trained on a dataset of humans walking on a treadmill and tested in real-time on overground walking. 
% Following previous studies \cite{}, we expect that overground walking may exhibit different kinematic characteristics compared to treadmill walking. Primarily because, during treadmill walking, parallel bars are used for balance, whereas overground walking involves the use of crutches. This second instrument promotes more unmeasured lateral movement in the exoskeleton. Additionally, in overground walking, subjects spontaneously choose their speed, while on the treadmill, they must adhere to a constant speed. These factors could lead to a decrease in the precision of the proposed method.
% Thus, in this second experiment we aim to evaluate to which extent the proposed model is able to generalize to movements not performed on the treadmill.

\section{Results}

\subsection{Validation of Deep-Learning Predictions}

% removed for now (could not reproduced w/ X2 data)
%Figure \ref{fig:autoencoder-300ms} presents the visualization of the autoencoder results from the benchmark dataset, using a window of 300ms. The autoencoder is a deep learning technique that projects the data in a lower dimension (here 3D) and then reconstructs the data to match the original input. We use this technique as a non-linear dimensionality reduction, to visualize the data. This Figure shows correlation between the kinematic window and the gait state, a suggestion that joint kinematics can be used to estimate the interpolation factor.

\begin{figure}[t] 
    \centering
    \vspace{0.2cm}
    \includegraphics[width=0.99\linewidth]{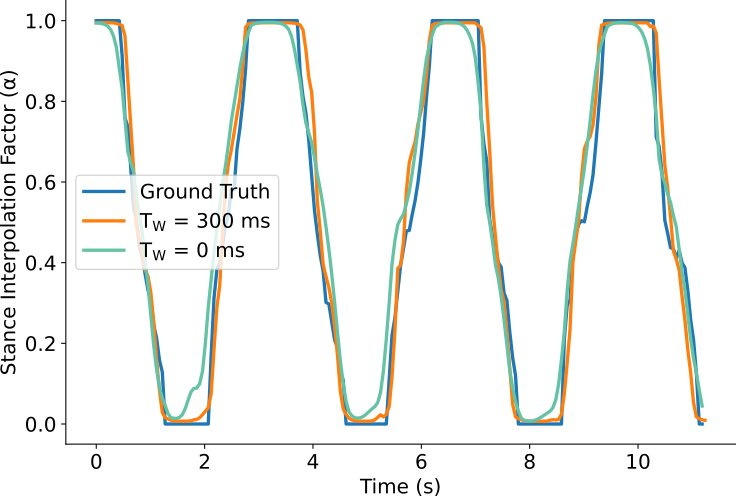}
    \caption{\small Validation of stance interpolation factor predictions for a representative user (US$_4$), using a window of 300~ms (orange) and with only the instantaneous joint configuration (green). Force plate condition (ground truth) is shown in blue.}
    \label{fig:results-offline-actual-history}
    \vspace{-0.2cm}
\end{figure}

%\begin{figure}[b]
%\centering
%\includegraphics[width=\linewidth]{gait-cycle2(1).png}
% \caption{\small Autoencoder with 300ms windows, with Right Gait cycle percentage. 22 users were included, walking at mainly 2 speeds 0.5 and 1.2 m/s, on a treadmill.}
%\label{fig:autoencoder-300ms}
%\end{figure}
% The proposed LSTM model, trained with offline data, underwent iterative training on data from four subjects and was subsequently tested on one excluded subject.
The proposed LSTM model was evaluated using a leave-one-out cross-validation approach: in each iteration, the model was trained on five users and then tested on the remaining user. In this test, we compared a model using instantaneous kinematic data ($T_W = 0$~ms) and another using a history of kinematic data ($T_W= 300$~ms).
Fig. \ref{fig:results-offline-actual-history} shows the prediction for each model evaluated on the test set composed by User Four and the ground truth value.

The model using instantaneous kinematic data resulted in a prediction time of $201 \pm 70.7$~\textmu s on a laptop (ThinkPad X1 Carbon 5th, Lenovo), while utilizing the history of kinematic data resulted in a prediction time of $573 \pm 308$~\textmu s. 

In terms of the prediction accuracy, the model incorporating a history of kinematic data produced a higher accuracy ($T_W = 0$~ms: $R^2 = 0.84 \pm 0.03$, MSE $ = 1.8 \times10^{-2} \pm 0.4$; $T_W= 300$~ms:  $R^2 = 0.90 \pm 0.02$, MSE $ =  1.1 \times10^{-2}\pm 0.3$). 
% Indeed,  the test $R^2$ value is $0.836 \pm 0.036$\footnote{Presented results represent the mean values derived from three users, accompanied by standard deviation information (mean ± SD).} using single input whereas using the window of kinematic data as input we obtain an $R^2$ of $0.902 \pm 0.027$ (MSE: $0.018 \pm 0.003 $ versus $ 0.011 \pm 0.003$).
% Indeed, in the single stance phase, the model with a 300ms window is able to predict a constant value close to 0 or 1, whereas using only the instantaneous position results in a prediction that is curved and does not achieve a steady-state value around 0 and 1. In addition, the double stance behavior is smoother and has fewer variations using more data in the input. %Indeed, the test $R^2$ value is 0.808 for user 4 data using single time input (MSE: 0.0196), whereas using the window of kinematic data as input we obtain an $R^2$ of 0.90 (MSE: 0.0102).
Tab. \ref{tab:results-offline-cross-user} provides a detailed report of the model's accuracy for individual users. Specifically, it includes MSE performances for both the training and test sets for $T_W = 0$ ms and $T_W = 300$ ms conditions.
\begin{table}
\center
\begin{tabular}{|c|c|c|c|c|c|}
\cline{3-6}
\multicolumn{2}{c|}{} & \multicolumn{2}{c|}{\textbf{$T_W$ = 0 ms}} & \multicolumn{2}{c|}{\textbf{$T_W$ = 300 ms}} \\ \hline
\textbf{Train-Set} & \textbf{Test-Set} & \textbf{Train} & \textbf{Test} & \textbf{Train} & \textbf{Test} \\ \hline
US$_{2,3,4,5,6}$ & US$_{1}$              & 0.0154   & 0.0199 & 0.0068 & 0.0146   \\ \hline
US$_{1, 3, 4,5,6}$ & US$_{2}$              & 0.0156  & 0.0182  & 0.0072 &  0.0109   \\ \hline
US$_{1, 2, 4,5,6}$ & US$_{3}$              & 0.0163  & 0.0133 & 0.0075    & 0.0082   \\ \hline
US$_{1, 2, 3,5,6}$ & US$_{4}$              & 0.0160  & 0.0176  & 0.0072    & 0.0125  \\ \hline
US$_{1, 2, 3,4,6}$ & US$_{5}$              & 0.0150  & 0.0234  & 0.0067    & 0.0123   \\ \hline
US$_{1, 2, 3,4,5}$ & US$_{6}$              & 0.0180  & 0.0126  & 0.0068    & 0.0081   \\ \hline
\end{tabular}
\caption{\small Prediction accuracy expressed as MSE for the LSTM model (for training and testing set) with time window of $300$~ms and instantaneous values. We performed a cross-user generalization evaluation by iteratively training on all users (US$_i$) except one and testing on the excluded user.}
\label{tab:results-offline-cross-user}
\vspace{-0.3cm}
\end{table}

% \begin{table}
% \begin{center}
% \caption{\small Prediction accuracy for the LSTM model ($R^2$ for training and testing set) with time-window of $300$~ms and instantaneous values. We performed a cross-user generalization evaluation by iteratively training on all users (US$_i$) except one and testing on the excluded user.}
% \label{tab:results-offline-cross-user}
% \begin{tabular}{|c|c|c|c|c|c|}
% \hline
% \multicolumn{2}{|c|}{} & \multicolumn{2}{c|}{\textbf{$R^2$ train}} & \multicolumn{2}{c|}{\textbf{$R^2$ test}} \\ \hline
% \multicolumn{2}{|c|}{} & \multicolumn{4}{c|}{\textbf{$T_W$}} \\ \hline
% \textbf{Train-Set} & \textbf{Test-Set} & \textbf{0 ms} & \textbf{300 ms} & \textbf{0 ms} & \textbf{300 ms} \\ \hline
% %\textbf{} & \textbf{} & \textbf{ms} & \textbf{ms} & \textbf{ms} & \textbf{ms} \\ \hline
% US$_{2,3,4}$ & US$_{1}$              & 0.87   & 0.94  &  0.81   & 0.88   \\ \hline
% US$_{1, 3, 4}$ & US$_{2}$              & 0.86  & 0.94  & 0.85    &  0.89   \\ \hline
% US$_{1, 2, 4}$ & US$_{3}$              & 0.86  & 0.93 &  0.88   & 0.94   \\ \hline
% US$_{1, 2, 3}$ & US$_{4}$              & 0.87  & 0.94  &  0.81  & 0.90   \\ \hline
% \end{tabular}
% \end{center}
% \end{table}

\subsection{Closed-Loop Performance During Treadmill Walking}

After validation of the deep-learning predictions, we compared the closed-loop performance of the proposed method for $\hat{\alpha}$ estimation and the measured ground truth $\alpha$ from the treadmill force plates. Specifically, estimated or measured stance interpolation factor values were used as inputs to the WECC controller~\cite{2024haptic} and resulting kinematics and interaction torque errors were evaluated during walking. During the closed-loop tests, we focused on the model with $T_W = 300$~ms due to its improved accuracy compared to $T_W = 0$~ms, while still obtaining a prediction time suitable for real-time implementation (0.57~ms).
% Specify closing the loop ...

\begin{figure*}[ht!]
        \centering 
        \vspace{0.2cm}%\includegraphics[width=\linewidth]{StanceInterp_Time_SD.png}
        \includegraphics[width=0.99\linewidth]{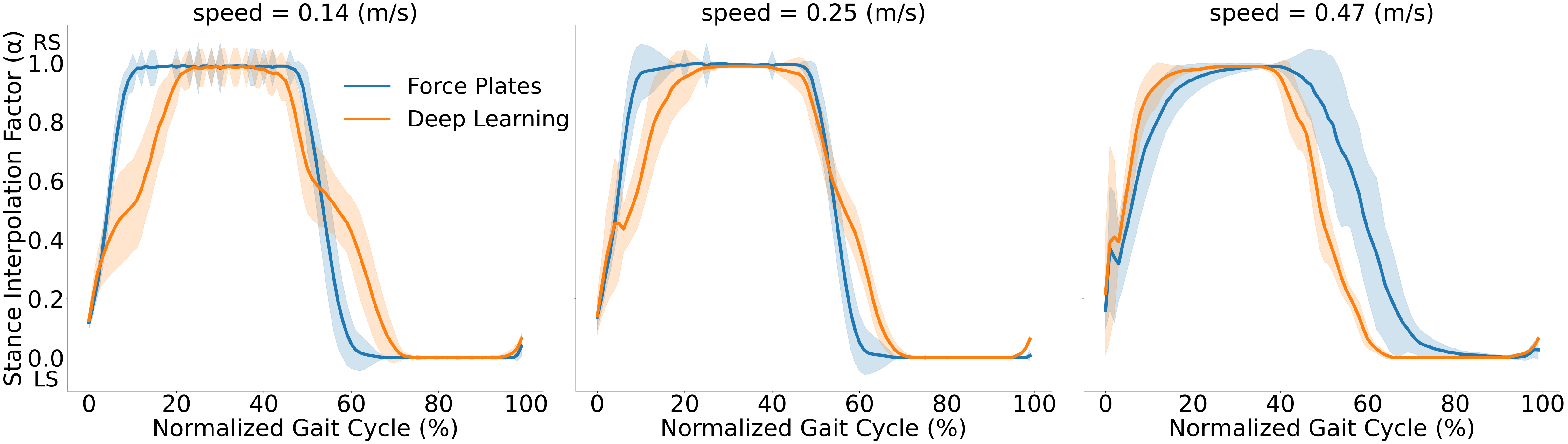}
        \caption[]%
        {{\small Stance interpolation factor using the treadmill force plates (blue) and using the deep-learning prediction (orange) for closed-loop exoskeleton control. Shaded error bars indicate $\pm$ one standard deviation relative to the mean. RS and LS denote Right Stance ($\alpha=1$) and Left Stance ($\alpha=0$), respectively. User walked at 0.14 m/s, 0.25 m/s and 0.47 m/s for one minute in every condition.}}    
        \label{fig:stance_interp_time_09}
        \vspace{-0.2cm}
\end{figure*}

In Fig. \ref{fig:stance_interp_time_09}, we illustrate the stance interpolation factor with respect to normalized gait duration under different conditions. The stance interpolation factor values obtained from treadmill force plates and the deep-learning approach are displayed for three different speeds (0.14 m/s, 0.25 m/s, 0.47 m/s). Compared to using force plate values, the proposed method resulted in longer stance time during slower walking and shorter stance time during faster walking. At 0.14 m/s, stance duration was $56.0 \pm 4.8$\% of the gait cycle using the treadmill force plates versus $67.1 \pm 9.7$\% using deep learning. At 0.25 m/s, stance duration was similar between the two approaches: $59.6 \pm 6.99$\% of the gait cycle using the treadmill force plates, and $65.1 \pm 1.6$\% using deep learning. Finally, at 0.47 m/s, the treadmill force plates resulted in a stance duration of $69.3 \pm 10.2$\% versus $59.7 \pm 5.74$\% using deep learning.

\begin{figure}[ht]
    \vspace{0.2cm}
    \centering 
    \includegraphics[width=\linewidth]{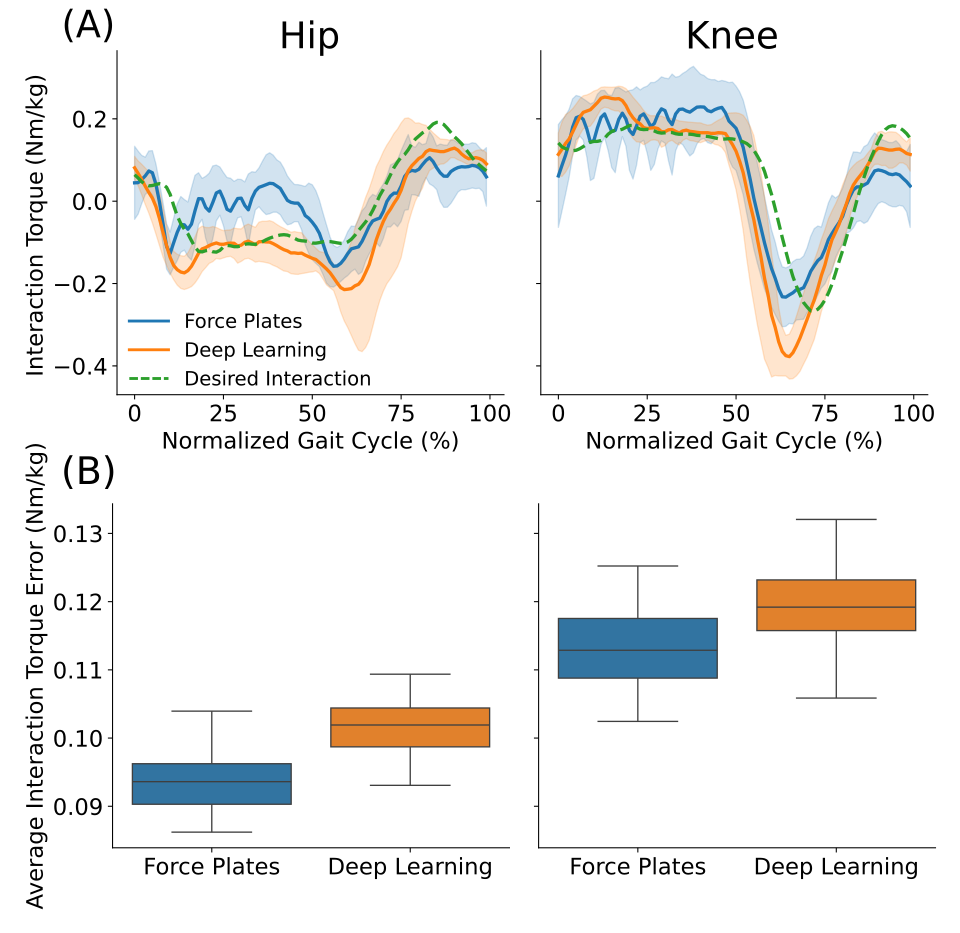}
    \caption[]%
    {{\small Haptic rendering performance of the proposed deep-learning method during treadmill walking. (A) Interaction torque across normalized gait cycle for a representative user, walking at 0.25 m/s. The interaction torque highlights haptic rendering performances using treadmill force plates (blue) or deep-learning estimation (orange). Shaded error bars indicate $\pm$ one standard deviation relative to the mean. Desired Interaction (in green) contains data from both treadmill force plates and deep-learning conditions. (B) Mean interaction torque error per step, for three users walking at 0.25 m/s. The boxplot shows highlights the interaction torque error in treadmill force plates and deep-learning conditions across users.}}   \label{fig:int_force_err_time_diff_haptic_rendering_boxplot}
    \vspace{-0.2cm}
\end{figure}

%Figures \ref{fig:int_force_err_time_diff_haptic_rendering_INT} and \ref{fig:int_force_err_time_diff_haptic_rendering_boxplot} show results of interaction forces rendered and errors of treadmill and machine learning conditions.  Actually, it is required that the system could render faithfully a spring-damper system to be used in future experiments. 
Interaction torque errors were investigated to evaluate the quality of the haptic rendering and transparency, using the force plates and the deep-learning predictions for the stance interpolation factor in real time. 
%Minimal interaction force errors would indicate enhanced transparency and haptic rendering performance. The system needs to demonstrate an accurate rendering of a spring-damper system to meet the requisites for future experiments.
Fig.~\ref{fig:int_force_err_time_diff_haptic_rendering_boxplot} presents the interaction torque tracking performance during the rendering mode while using the force plate measurements and deep-learning estimation in the dynamic robot model. It was observed that interaction torque error is higher with the deep-learning method at the start of the swing. Specifically, toe-off (60-70\% of the gait cycle) led to an average hip interaction torque error of $0.068$ Nm/kg using force plates and $0.109$ Nm/kg using deep learning. However, this error was lower for the deep-learning approach in the stance phase. Within the 20-50\% range of the gait cycle, the average hip interaction error was $0.111$ Nm/kg using force plates and $0.033$ Nm/kg using deep learning. The mean of the interaction torque error over the whole cycle, across three users, was comparable between the two conditions. At the hip joint, mean absolute interaction torque errors of $0.101 \pm 0.004$~Nm/kg and $0.094 \pm 0.005$~Nm/kg were observed with the deep-learning and force plate conditions, respectively. Similarly, the mean absolute interaction torque error at the knee joint was $0.119 \pm 0.006$~Nm/kg and $0.113 \pm 0.006$~Nm/kg for the deep-learning and force plate conditions.

% Additionally, the interaction force error is lower at the hip when using the deep learning estimation, with a measurement of $4.42 \pm 0.88$~Nm/kg, in contrast to $5.92 \pm 2.27$~Nm/kg when relying on force plates. At the knee joint, the interaction force error is comparable between the two methods with values of $5.45 \pm 1.87$~Nm/kg for deep learning and $5.47 \pm 2.06$~Nm/kg for force plate measurements.

\begin{figure}[ht!] 
    \vspace{0.2cm}
    \centering 
    \includegraphics[width=\linewidth]{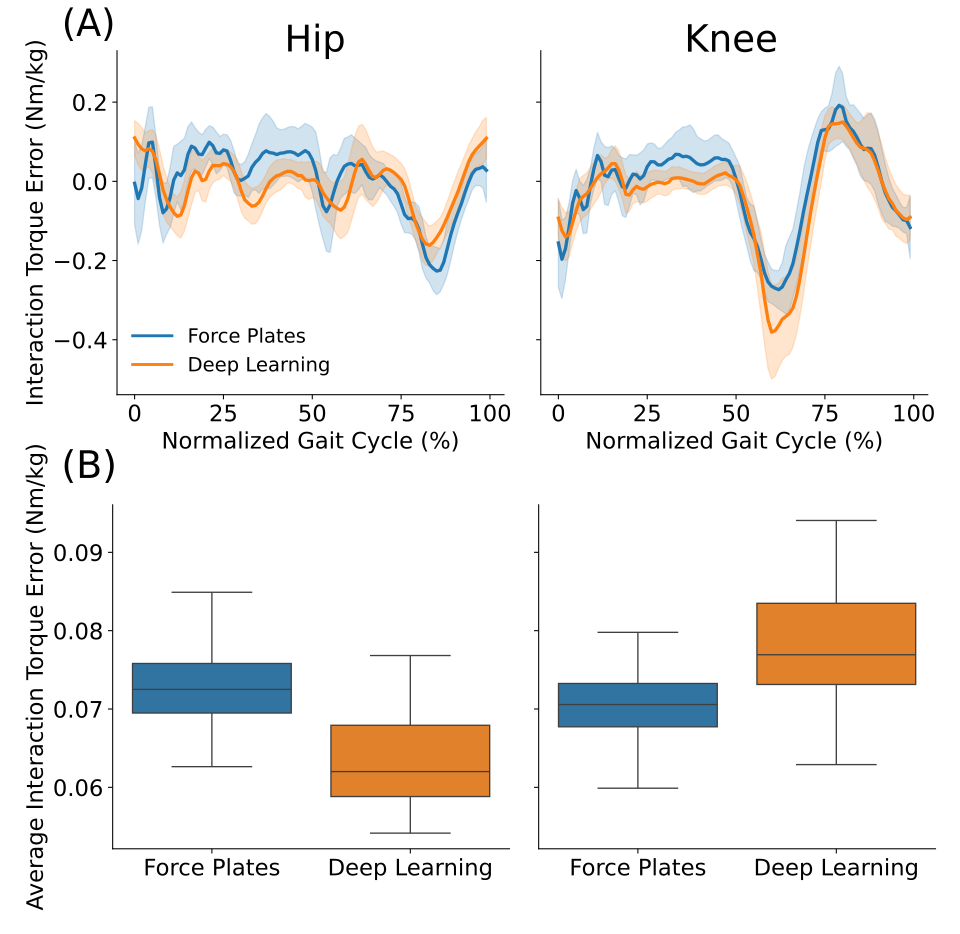}
    \caption[]%
    {{\small Haptic transparency performance of the proposed deep-learning method during treadmill walking (desired interaction torque is zero). (A) Interaction torque error with respect to normalized gait cycle for a representative user walking at 0.25 m/s. Shaded error bars indicate $\pm$ one standard deviation relative to the mean. (B) Boxplot of mean interaction torque error over each step, at 0.25 m/s. Results are aggregated over three users.}} \label{fig:int_force_err_boxplot_3_users_avg}
    \vspace{-0.2cm}
\end{figure}

Fig. \ref{fig:int_force_err_boxplot_3_users_avg} shows the results of the transparency performances on both conditions. The deep-learning approach resulted in a relatively smaller hip interaction torque error compared to the force-plate condition ($0.063 \pm 0.006$~Nm/kg vs. $0.073 \pm 0.008$~Nm/kg). On the other hand, the deep-learning approach resulted in a higher interaction torque error at the knee joints between 60\% and 70\% of the normalized gait, corresponding to the beginning of the swing phase. The mean knee interaction torque error in this period was $-0.207$~Nm/kg for the force plate condition and $-0.312$~Nm/kg for the deep-learning approach. Users also qualitatively reported that the exoskeleton felt heavier at this particular moment of the gait cycle. Across the whole gait cycle, the mean interaction torque error for the knee joint was similar for the deep-learning and the force plate conditions ($0.078 \pm 0.007$~Nm/kg vs. $0.071 \pm 0.007$~Nm/kg).

% \begin{figure}[ht!]
%     \centering 
%     \includegraphics[width=\linewidth]{new_version/HAPTIC_IntForce_Time_SD_GoodCycle_haptic_09_4joints_one_user.png}
%     \caption[]%
%     {{\small Interaction Force across normalized gait cycle, averaged over three users, at 0.25 m/s. The interaction force highlights haptic rendering performances using treadmill (blue) or online machine learning (orange). Lighter bands show Standard Deviation. Desired Interaction (in green) contains data from both treadmill and machine learning conditions.}}    
%     \label{fig:int_force_err_time_diff_haptic_rendering_INT}
% \end{figure}
  
% \begin{figure}[ht!]
%     \centering 
%     \includegraphics[width=\linewidth]{new_version/demeaned_boxplot_avg_int_force_haptic.png}
%     \caption[]%
%     {{\small Average Interaction Force Error per step, for three users at 0.25 m/s. The boxplot shows haptic rendering performances, highlighting the interaction force error in treadmill and machine learning conditions.}}   \label{fig:int_force_err_time_diff_haptic_rendering_boxplot}
% \end{figure}

% \begin{figure}[ht!]
%     \centering 
%     \includegraphics[width=\linewidth]{new_version/IntForce_Time_SD_GoodCycle_haptic_09_4joints_one_user(1).png}
%     \caption[Network2]%
%     {{\small Interaction Force Error with respect to normalized gait cycle. The data from a representative user at 0.25 m/s is shown. The standard deviation is displayed on the lighter bands to show the variability of both signals.} 
%     }   
%     \label{fig:int_force_err_time_3users}
% \end{figure}

% \begin{figure}[ht!] 
%     \centering 
%     \includegraphics[width=\linewidth]{new_version/demeaned_boxplot_avg_int_force_transp.png}
%     \caption[]%
%     {{\small Boxplot of Average Interaction Force Error over each step, at 0.25 m/s. Results are aggregated over three users.}}    
%     \label{fig:int_force_err_boxplot_3_users_avg}
% \end{figure}

\subsection{Closed-Loop Performance During Overground Walking}

\begin{figure}[t] 
    \centering 
    \vspace{0.2cm}
    \includegraphics[width=\linewidth]{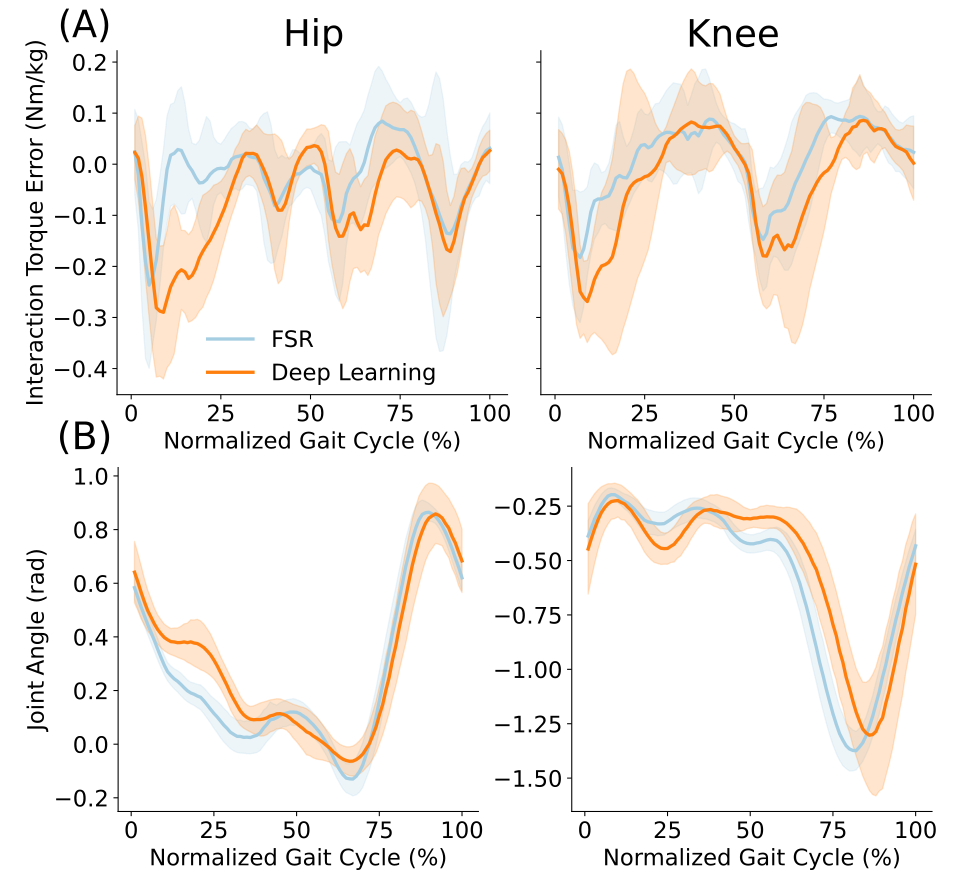}
    \caption[]%
    {{\small Haptic transparency performance of the proposed deep-learning method during overground walking. (A) Interaction torque across normalized gait cycle, for a representative user, during overground walking. (B) Hip and Knee joint angles obtained with deep learning (orange) and FSR pad sensors (blue) conditions for a representative user. Shaded error bars indicate $\pm$ one standard deviation relative to the mean.}}    
\label{fig:torque_error_time_one_user_lvg}
\vspace{-0.2cm}
\end{figure}

%Figure \ref{fig:step_time_avg_2_users} shows the selected step duration and resultant joint kinematics during overground walking tests at self-selected walking pace. 
%It was observed that the users exhibited a similar step duration using the FSR footplates with $ 1.16 \pm 0.06$~s compared to the deep-learning online estimation with $1.28 \pm 0.10$~s. 
During the overground tests, it was observed that the users preferred to walk at a slightly faster speed using the FSR footplates ($0.241 \pm 0.024$~m/s) compared to the deep-learning estimation ($0.198 \pm 0.020$~m/s). Results from Fig.~\ref{fig:torque_error_time_one_user_lvg}B show similar kinematic patterns between the two conditions, particularly in the magnitudes of the joint angles. In the temporal domain, a longer stance period was observed with the deep-learning approach ($75.3 \pm 2.7$\%) compared to the FSR condition ($63.1 \pm 1.6$\%). % as shown in Figure~\ref{fig:interpol_avg_one_user_lvg}.
Furthermore, Fig.~\ref{fig:torque_error_time_one_user_lvg} presents the interaction torque error over a gait cycle, for a representative user, using FSRs or deep learning. The interaction torque error, averaged over two users, was higher for the deep-learning condition (Hip: $0.094 \pm 0.009$ Nm/kg using FSR and $0.120 \pm  0.014$ Nm/kg using deep learning;  Knee: $0.083 \pm 0.007$ Nm/kg using FSR and $0.116 \pm 0.009$ Nm/kg using deep learning).

% \begin{figure}[h!] 
%     \centering 
%     \includegraphics[width=\linewidth]{new_version/Hip_angle_2_users_all_data.png}
%     \caption[]%
%     {{\small Hip angles obtained with machine learning (orange) and FSR pad sensors (blue) conditions. Results are aggregated over two users.}}    
%     \label{fig:hip_avg_2_users}
% \end{figure}

% \begin{figure}[h!] 
%     \centering 
%     \includegraphics[width=\linewidth]{new_version/Knee_angle_2_users_all_data.png}
%     \caption[]%
%     {{\small Hip angles obtained with machine learning (orange) and FSR pad sensors (blue) conditions. Results are aggregated over two users.}}    
%     \label{fig:knee_avg_2_users}
% \end{figure}

\section{Discussion}

%%%%%%%%%%%%%%%%%%%%%%%%%%%%%%%%%
% structure:
%%%%%%%%%%%%%%%%%%%%%%%%%%%%%%%%%%%
% offline performance:
% - prediction time 
% - prediction accuracy: discuss fig 2 (steady state on 0 and 1 for 300ms model, model generability and not overfitting in table 1)

In this paper, we evaluated a deep-learning model to predict the weight distribution (i.e., stance interpolation factor) of a user wearing a lower-limb exoskeleton during several walking conditions; we compared the exoskeleton controller performance when this stance interpolation factor was estimated with our deep-learning model ($\hat{\alpha}$) and measured with ground truth values (treadmill force plates or FSR sensor pads, $\alpha$). This work highlights the viability and limitations of using deep-learning predictions to detect changes in the gait state using only joint kinematics information without ground reaction sensors.

% Here comes a paragraph on using window of kinematic val.
Utilizing a history of kinematic data was shown to enhance the model's accuracy (Fig. \ref{fig:results-offline-actual-history} and Tab. \ref{tab:results-offline-cross-user}) in comparison to using instantaneous values, as seen in a similar study \cite{icinco21}. These findings show the advantage of utilizing additional temporal data from previous time steps, enhancing the network's performance in both training and testing phases. Importantly, including additional data did not significantly alter the model's prediction speed; we observed comparable prediction times when using a history of kinematic data with respect to the instantaneous value predictions.

As a result, the model was implemented to achieve real-time performance. This is crucial for its use in the control of an exoskeleton, as predictions need to be faster than the main control loop. Achieving an average prediction time of 0.57~ms ensured the real-time usability of the system, as the main control loop runs usually significantly slower than 1~ms on lower-limb exoskeletons (e.g., 3~ms for the exoskeleton controller in our study). It is noteworthy that, without using TensorFlow Lite (i.e., employing the classic TensorFlow library), the average prediction time with a history of kinematic data is $67.1 \pm 13.5$~ms, which cannot be used in most real-time applications. Furthermore, the proposed method provides accurate predictions of the stance interpolation factor for previous and unseen users (Fig. \ref{fig:results-offline-actual-history} and Tab. \ref{tab:results-offline-cross-user}), demonstrating another aspect of its usability. Requiring training data for every exoskeleton user is not convenient or realistic, particularly in physical rehabilitation as a patient's time spent receiving therapy must be prioritized.

In the context of machine learning for robotic control, using the predictions in a closed loop can affect the performance due to error propagation in real time \cite{levine2020offline}. Therefore, it was critical to evaluate the closed-loop performance of our system, using several conditions. We demonstrated that the deep-learning approach produces a similar stance interpolation factor compared to treadmill force plates during walking with haptic transparent control; these findings generalize for walking speeds between 0.14 and 0.47 m/s (Fig. \ref{fig:stance_interp_time_09}). To characterize the effect of the deep-learning predictions on the performance of the exoskeleton controller, we assessed both haptic rendering of nonzero impedance and haptic transparent modes during treadmill walking. Moreover, we evaluated the extent to which these results generalize to overground walking. During treadmill walking, mean interaction torque errors were similar between deep-learning predictions and treadmill force plate measurements. Evaluating the performance of the controller during more naturalistic, overground walking, the deep-learning predictions also facilitated similar interaction torque error profiles (Fig. \ref{fig:torque_error_time_one_user_lvg}A) and hip and knee kinematics (Fig. \ref{fig:torque_error_time_one_user_lvg}B).

While the overall performance of the exoskeleton controller was comparable when implementing our deep-learning predictions and ground reaction sensor values, we did observe a delay-like effect in the stance interpolation factor estimation which likely resulted in some performance discrepancies at specific phases of the gait cycle. Specifically, we observed a longer stance duration when walking with the deep-learning predictions at slower speeds (Fig. \ref{fig:stance_interp_time_09}); this was associated with a longer transition between left and right single stance (i.e., double stance to swing).

During treadmill and overground walking, at the knee joint, we observed higher interaction torque error at the beginning of the swing phase (around 60-70\% of the cycle, Fig.~\ref{fig:int_force_err_boxplot_3_users_avg}A) with the deep-learning predictions. This increase in interaction torque error was more noticeable during overground walking, and also observed at the hip joint. In this condition, the use of crutches may have promoted additional lateral movement which was not measured by the exoskeleton \cite{Alamro2018}. 
% This increase could be attributed to the prolonged transition during double stance (Fig. \ref{fig:stance_interp_time_09}), requiring additional torque from the user when initiating the swing phase.

For both treadmill and overground walking, the discrepancies in interaction torque errors relative to their respective ground truths could be due to the deep-learning model requiring users to change their kinematics before the stance interpolation factor changes. This is different from using force sensors, which can detect changes in weight distribution independent of kinematics recorded by the exoskeleton. One way to improve the detection of changes in weight distribution in the deep-learning model is to incorporate the frontal plane IMU angle of the backpack in addition to the sagittal plane angles from the backpack and joint encoders.

\section{Conclusion}

In this study, we demonstrated the feasibility of employing deep learning in gait state detection for exoskeleton control, eliminating the necessity of ground reaction force sensors. Despite challenges such as a limited dataset, real-time constraints, and issues related to error propagation, these obstacles were successfully addressed. Evaluation of our model's closed-loop performance versus typical sensors (i.e., force plates, FSR footplates) highlights the system's applicability across new users and several walking speeds and conditions. To address our model limitations, future work could include training datasets encompassing various walking speeds, overground walking and other activities (e.g. ramps, stairs).
Furthermore, this work could be used to implement a state machine that utilizes an estimated stance interpolation factor, selecting impedance parameters in the mid-level controller for exoskeleton control and tailored assistance during walking. The proposed framework will be validated in future on additional healthy individuals as well as individuals with lower-limb impairments (e.g., stroke, spinal cord injury).

\addtolength{\textheight}{-12cm}   % This command serves to balance the column lengths
                                  % on the last page of the document manually. It shortens
                                  % the textheight of the last page by a suitable amount.
                                  % This command does not take effect until the next page
                                  % so it should come on the page before the last. Make
                                  % sure that you do not shorten the textheight too much.

%%%%%%%%%%%%%%%%%%%%%%%%%%%%%%%%%%%%%%%%%%%%%%%%%%%%%%%%%%%%%%%%%%%%%%%%%%%%%%%%

%%%%%%%%%%%%%%%%%%%%%%%%%%%%%%%%%%%%%%%%%%%%%%%%%%%%%%%%%%%%%%%%%%%%%%%%%%%%%%%%

%%%%%%%%%%%%%%%%%%%%%%%%%%%%%%%%%%%%%%%%%%%%%%%%%%%%%%%%%%%%%%%%%%%%%%%%%%%%%%%%
% \section*{APPENDIX}

% Appendixes should appear before the acknowledgment.

\section*{Acknowledgment}

This work was supported by the National Science Foundation~/~National Robotics Initiative (Grant No: 2024488).  We would like to thank Tim Haswell for his technical support on the hardware improvements of the ExoMotus-X2 exoskeleton.

\bibliographystyle{IEEEtran}
\bibliography{ref_partic}

% Generated by IEEEtran.bst, version: 1.14 (2015/08/26)
\begin{thebibliography}{10}
\providecommand{\url}[1]{#1}
\csname url@samestyle\endcsname
\providecommand{\newblock}{\relax}
\providecommand{\bibinfo}[2]{#2}
\providecommand{\BIBentrySTDinterwordspacing}{\spaceskip=0pt\relax}
\providecommand{\BIBentryALTinterwordstretchfactor}{4}
\providecommand{\BIBentryALTinterwordspacing}{\spaceskip=\fontdimen2\font plus
\BIBentryALTinterwordstretchfactor\fontdimen3\font minus \fontdimen4\font\relax}
\providecommand{\BIBforeignlanguage}[2]{{%
\expandafter\ifx\csname l@#1\endcsname\relax
\typeout{** WARNING: IEEEtran.bst: No hyphenation pattern has been}%
\typeout{** loaded for the language `#1'. Using the pattern for}%
\typeout{** the default language instead.}%
\else
\language=\csname l@#1\endcsname
\fi
#2}}
\providecommand{\BIBdecl}{\relax}
\BIBdecl

\bibitem{Baud2021}
\BIBentryALTinterwordspacing
R.~Baud, A.~R. Manzoori, A.~Ijspeert, and M.~Bouri, ``Review of control strategies for lower-limb exoskeletons to assist gait,'' \emph{Journal of {NeuroEngineering} and Rehabilitation}, vol.~18, no.~1, Jul. 2021. [Online]. Available: \url{https://doi.org/10.1186/s12984-021-00906-3}
\BIBentrySTDinterwordspacing

\bibitem{deMiguelFernndez2023}
J.~de~Miguel-Fernández, J.~Lobo-Prat, E.~Prinsen, J.~M. Font-Llagunes, and L.~Marchal-Crespo, ``Control strategies used in lower limb exoskeletons for gait rehabilitation after brain injury: a systematic review and analysis of clinical effectiveness,'' \emph{Journal of NeuroEngineering and Rehabilitation}, vol.~20, no.~1, Feb. 2023.

\bibitem{9513580}
E.~Basalp, P.~Wolf, and L.~Marchal-Crespo, ``Haptic training: Which types facilitate (re)learning of which motor task and for whom? answers by a review,'' \emph{IEEE Transactions on Haptics}, vol.~14, no.~4, pp. 722--739, 2021.

\bibitem{Kim2022}
M.~Kim, A.~M. Simon, and L.~J. Hargrove, ``Seamless and intuitive control of a powered prosthetic leg using deep neural network for transfemoral amputees,'' \emph{Wearable Technologies}, vol.~3, 2022.

\bibitem{8386443}
W.~Huo, S.~Mohammed, Y.~Amirat, and K.~Kong, ``Fast gait mode detection and assistive torque control of an exoskeletal robotic orthosis for walking assistance,'' \emph{IEEE Transactions on Robotics}, vol.~34, no.~4, pp. 1035--1052, 2018.

\bibitem{8487913}
A.~Ortlieb, R.~Baud, T.~Tracchia, B.~Denkinger, Q.~Herzig, H.~Bleuler, and M.~Bouri, ``An active impedance controller to assist gait in people with neuromuscular diseases: Implementation to the hip joint of the autonomyo exoskeleton,'' in \emph{2018 7th IEEE International Conference on Biomedical Robotics and Biomechatronics (Biorob)}, 2018.

\bibitem{2024haptic}
E.~B. Küçüktabak, Y.~Wen, S.~J. Kim, M.~R. Short, D.~Ludvig, L.~Hargrove, E.~J. Perreault, K.~M. Lynch, and J.~L. Pons, ``Haptic transparency and interaction force control for a lower-limb exoskeleton,'' \emph{IEEE Transactions on Robotics}, pp. 1--19, 2024.

\bibitem{MUNDT202029}
M.~Mundt, A.~Koeppe, S.~David, F.~Bamer, W.~Potthast, and B.~Markert, ``Prediction of ground reaction force and joint moments based on optical motion capture data during gait,'' \emph{Medical Engineering \& Physics}, vol.~86, pp. 29--34, 2020.

\bibitem{Karatsidis2016}
\BIBentryALTinterwordspacing
A.~Karatsidis, G.~Bellusci, H.~Schepers, M.~de~Zee, M.~Andersen, and P.~Veltink, ``Estimation of ground reaction forces and moments during gait using only inertial motion capture,'' \emph{Sensors}, vol.~17, no.~12, p.~75, Dec. 2016. [Online]. Available: \url{https://doi.org/10.3390/s17010075}
\BIBentrySTDinterwordspacing

\bibitem{https://doi.org/10.1111/cgf.14635}
L.~Mourot, L.~Hoyet, F.~L. Clerc, and P.~Hellier, ``Underpressure: Deep learning for foot contact detection, ground reaction force estimation and footskate cleanup,'' \emph{Computer Graphics Forum}, vol.~41, no.~8, pp. 195--206, 2022.

\bibitem{s151127738}
J.-Y. Jung, W.~Heo, H.~Yang, and H.~Park, ``A neural network-based gait phase classification method using sensors equipped on lower limb exoskeleton robots,'' \emph{Sensors}, vol.~15, no.~11, pp. 27\,738--27\,759, 2015.

\bibitem{Liu2016}
D.-X. Liu, X.~Wu, W.~Du, C.~Wang, and T.~Xu, ``Gait phase recognition for lower-limb exoskeleton with only joint angular sensors,'' \emph{Sensors}, vol.~16, no.~10, p. 1579, Sep. 2016.

\bibitem{9415159}
C.~Camardella, F.~Porcini, A.~Filippeschi, S.~Marcheschi, M.~Solazzi, and A.~Frisoli, ``Gait phases blended control for enhancing transparency on lower-limb exoskeletons,'' \emph{IEEE Robotics and Automation Letters}, vol.~6, no.~3, pp. 5453--5460, 2021.

\bibitem{icinco21}
V.~Lippi, C.~Camardella, A.~Filippeschi, and F.~Porcini, ``Identification of gait phases with neural networks for smooth transparent control of a lower limb exoskeleton,'' in \emph{Proceedings of the 18th International Conference on Informatics in Control, Automation and Robotics - Volume 1: ICINCO,}, INSTICC.\hskip 1em plus 0.5em minus 0.4em\relax SciTePress, 2021, pp. 171--178.

\bibitem{10.1007/978-3-030-69547-7_47}
J.~Fong, E.~B. K{\"u}{\c{c}}{\"u}ktabak, V.~Crocher, Y.~Tan, K.~M. Lynch, J.~L. Pons, and D.~Oetomo, ``Canopen robot controller (corc): An open software stack for human robot interaction development,'' in \emph{Wearable Robotics: Challenges and Trends}, J.~C. Moreno, J.~Masood, U.~Schneider, C.~Maufroy, and J.~L. Pons, Eds.\hskip 1em plus 0.5em minus 0.4em\relax Cham: Springer International Publishing, 2022, pp. 287--292.

\bibitem{s20216345}
F.~Labarrière, E.~Thomas, L.~Calistri, V.~Optasanu, M.~Gueugnon, P.~Ornetti, and D.~Laroche, ``Machine learning approaches for activity recognition and/or activity prediction in locomotion assistive devices—a systematic review,'' \emph{Sensors}, vol.~20, no.~21, 2020.

\bibitem{9842329}
M.~Kim and L.~J. Hargrove, ``Deep-learning to map a benchmark dataset of non-amputee ambulation for controlling an open source bionic leg,'' \emph{IEEE Robotics and Automation Letters}, vol.~7, no.~4, pp. 10\,597--10\,604, 2022.

\bibitem{kingma2014adam}
D.~P. Kingma and J.~Ba, ``Adam: A method for stochastic optimization,'' \emph{arXiv preprint arXiv:1412.6980}, 2014.

\bibitem{Kucuktabak2023virtual}
E.~B. Küçüktabak, Y.~Wen, M.~Short, E.~Demirbaş, K.~Lynch, and J.~Pons, ``Virtual physical coupling of two lower-limb exoskeletons,'' in \emph{2023 International Conference on Rehabilitation Robotics (ICORR)}, 2023, pp. 1--6.

\bibitem{levine2020offline}
S.~Levine, A.~Kumar, G.~Tucker, and J.~Fu, ``Offline reinforcement learning: Tutorial, review, and perspectives on open problems,'' \emph{arXiv preprint arXiv:2005.01643}, 2020.

\bibitem{Alamro2018}
R.~A. Alamro, A.~E. Chisholm, A.~M.~M. Williams, M.~G. Carpenter, and T.~Lam, ``Overground walking with a robotic exoskeleton elicits trunk muscle activity in people with high-thoracic motor-complete spinal cord injury,'' \emph{Journal of NeuroEngineering and Rehabilitation}, vol.~15, no.~1, Nov. 2018.

\end{thebibliography}

\end{document}